\documentclass{article}

\usepackage{arxiv}

\usepackage[utf8]{inputenc}
\usepackage[T1]{fontenc}
\usepackage{url}
\usepackage{booktabs}
\usepackage{amsmath}
\usepackage{amssymb}
\usepackage{amsfonts}
\usepackage{amsthm}
\usepackage{multirow}
\usepackage{mathtools}
\usepackage{nicefrac}
\usepackage{microtype}
\usepackage[table]{xcolor}
\usepackage{graphicx}
\usepackage{bm}
\usepackage{adjustbox}
\usepackage{tikz}
\usepackage{wrapfig}
\usepackage{subcaption}
\usepackage{enumitem}
\usepackage{svg}
\usepackage{bbm}

\usetikzlibrary{arrows.meta,positioning,fit,calc}

\newcommand{\best}[1]{\textbf{#1}}

\newtheorem{theorem}{Theorem}
\newtheorem{lemma}{Lemma}

\renewcommand{\E}{\mathbb{E}}
\renewcommand{\Var}{\operatorname{Var}}
\newcommand{\KL}{\operatorname{KL}}

\newcommand{\pitheta}{\pi_\theta}
\newcommand{\piref}{\pi_{\mathrm{ref}}}

\newcommand{\dbar}{\bar\delta}
\newcommand{\dhat}{\widehat\delta}
\newcommand{\Atilde}{\widetilde A}

\newcommand{\sg}{\operatorname{sg}}
\newcommand{\ellRSPO}{\ell_{\mathrm{RSPO}}}
\newcommand{\ellAW}{\ell_{\mathrm{AW}}}

\newcommand{\VarG}{\operatorname{Var}_{\mathcal B}}
\newcommand{\innerG}[2]{\langle #1,#2\rangle_{\mathcal B}}
\newcommand{\normG}[1]{\|#1\|_{\mathcal B}}

\title{Relative Score Policy Optimization for Diffusion Language Models}

\author{
  Zichao Yu$^{1}$\thanks{These authors contributed equally to this work.}\quad
  Shengze Xu$^{2}$\footnotemark[1]\quad
  Bingqing Jiang$^{3}$\quad
  Wenyi Zhang$^{1}$\quad
  Difan Zou$^{3}$\thanks{Corresponding author}
  \\
  \\
  $^{1}$University of Science and Technology of China\quad
  $^{2}$The Chinese University of Hong Kong\\
  $^{3}$The University of Hong Kong
}
\date{}

\begin{document}

\maketitle

\begin{abstract}
Diffusion large language models (dLLMs) offer a promising route to parallel and efficient text generation, but improving their reasoning ability requires effective post-training. Reinforcement learning with verifiable rewards (RLVR) is a natural choice for this purpose, yet its application to dLLMs is hindered by the absence of tractable sequence-level log-ratios, which are central to standard policy optimization. 
The lack of tractable sequence-level log-ratios forces existing methods to rely
on high-variance ELBO-based approximations, where high verifier rewards can
amplify inaccurate score estimates and destabilize RL training. To overcome this issue, we propose \textbf{R}elative \textbf{S}core \textbf{P}olicy \textbf{O}ptimization (RSPO), a simple RLVR method that uses verifiable rewards to calibrate noisy likelihood estimates in dLLMs. 
The core of our algorithm relies on a key observation: a reward advantage can be interpreted not only as an
update direction, but also as a target for the relative log-ratio between the
current and reference policies. 
Accordingly, RSPO calibrates this noisy relative log-ratio estimate by comparing its reward advantage with the reward-implied target relative log-ratio, updating the policy according to the gap between the current estimate and the target rather than the raw advantage alone. 
Experiments on mathematical reasoning and planning benchmarks show that RSPO yields especially strong gains on planning tasks and competitive mathematical-reasoning performance.
\end{abstract}

\section{Introduction}
Reinforcement learning with verifiable rewards (RLVR)~\citep{he2025response,shao2024deepseekmath,yu2025dapo,zheng2025group} has become an effective post-training paradigm for improving language models on reasoning tasks: sample candidate answers, verify their correctness, and update the model so that correct answers become more likely~\citep{sutton1998reinforcement,ouyang2022training,jaech2024openai,guo2025deepseek}. 
For autoregressive language models, this update is naturally expressed through policy log-ratios, namely the change in log-likelihood of a sampled response under the current policy relative to a reference or previous policy. 
Because autoregressive models admit an explicit left-to-right factorization, these log-ratios are tractable, enabling methods such as Proximal Policy Optimization (PPO)~\citep{schulman2017proximal} and Group Relative Policy Optimization (GRPO)~\citep{shao2024deepseekmath} to combine reward improvement with Kullback--Leibler (KL) control.

Diffusion large language models (dLLMs)~\citep{sahoo2024simple,shi2024simplified,nielarge,yang2025mmada,bie2025llada2} break this convenient interface. Their generation process proceeds through iterative denoising rather than left-to-right prediction, so the sequence likelihood of a completed response is not directly available. 
As a result, the sequence-level policy log-ratios used by standard RL objectives become difficult to compute.
Recent RL methods for dLLMs therefore replace exact log-ratios with tractable surrogates, such as one-step mean-field estimates~\citep{zhao2025d1}, ELBO-based token or sequence scores~\citep{ou2025principled}, trajectory-level formulations~\citep{wang2025revolutionizing}, and evidence-bound objectives~\citep{wang2025spg,lin2025boundary}. 
These substitutes make policy optimization possible, but they also introduce a new source of uncertainty: in RLVR, the verifier reward is a reliable task signal, whereas the model-side relative score used to apply that reward is only an approximate and noisy estimate of the true likelihood change.

\begin{figure}[t]
    \centering
\includegraphics[width=0.8\linewidth]{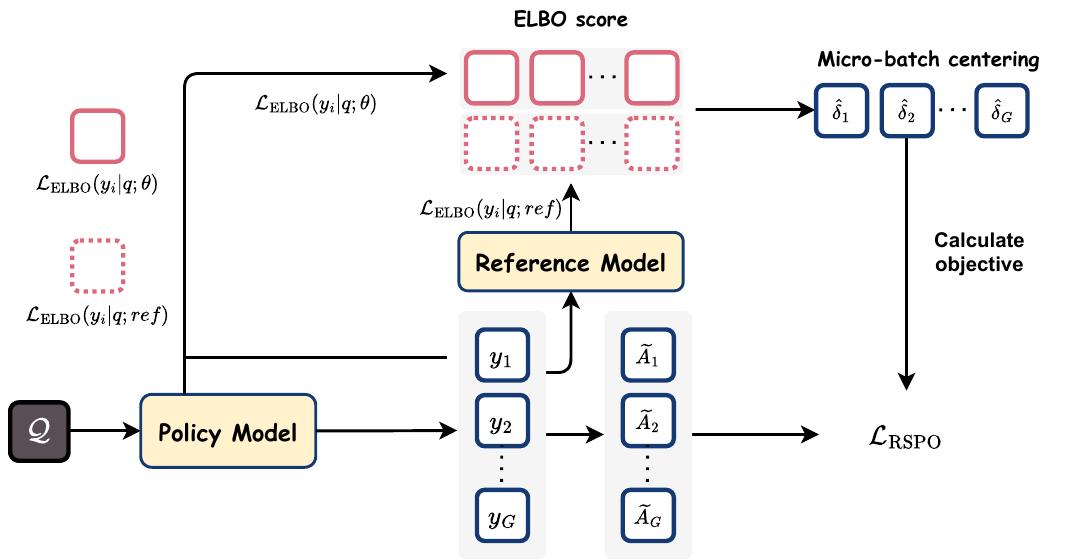}
    \caption{\small Overview of RSPO: verifier rewards define centered relative-score targets, and the residual to the current ELBO-based score provides policy-update feedback.}
    \label{fig:RSPO}
\end{figure}
This mismatch raises a basic question: how should dLLM RLVR combine a reliable verifier reward with a noisy likelihood-ratio surrogate? Standard advantage-weighted objectives address only part of this question. They multiply a model-side score by a reward or advantage coefficient, thereby specifying which responses should be encouraged. When the score is an ELBO-based log-ratio surrogate, however, the coefficient does not depend on the current relative score. A high-advantage response can therefore continue to amplify a noisy likelihood estimate even after its estimated relative likelihood has moved far from the reference. Thus, the reward provides a preference direction, but the objective lacks a calibrated target for the relative score itself.

Our key observation is that KL-regularized policy improvement provides exactly this target. 
The optimal KL-regularized policy assigns each response a current-reference log-ratio determined by its reward, up to a prompt-level normalization term. 
Within a comparison group, this normalization term is shared and disappears after centering. 
The centered verifier rewards therefore specify the relative likelihood changes that the improved policy should assign to sampled responses. 
Based on this principle, we propose \textbf{R}elative \textbf{S}core \textbf{P}olicy \textbf{O}ptimization (RSPO), a simple RLVR objective for diffusion language models. 
RSPO estimates a centered current-reference score with ELBO differences, derives a centered relative-score target from verifier rewards, and uses the residual gap as feedback in the policy update. 
In this way, reliable rewards calibrate noisy likelihood-ratio estimates instead of serving only as fixed multiplicative weights. 
Figure~\ref{fig:RSPO} gives an overview of this feedback loop: the policy generates a group of responses for each prompt, the reference model is used to form ELBO-based log-ratio scores, and micro-batch centering turns these scores into relative feedback. RSPO compares this current centered score with the verifier-induced target and optimizes the resulting residual objective; details are deferred to Section~\ref{sec:method}.
Experiments on mathematical reasoning and planning benchmarks show that RSPO yields large planning gains and competitive mathematical-reasoning performance against existing RL baselines.

Our contributions are summarized as follows:
\begin{itemize}[leftmargin=*]
    \item We identify a calibration problem in dLLM RLVR. Verifier rewards provide reliable task-level feedback, but the ELBO-based log-ratios used for policy optimization are noisy likelihood-ratio surrogates. Standard advantage-weighted objectives use these rewards as fixed multiplicative coefficients, which can keep pushing already high-scoring responses and over-amplify noisy relative-score estimates.
    \item We derive a principled relative-score target from KL-regularized policy improvement. Although the optimal policy log-ratio contains an unknown normalization term, this term is shared within a comparison group and can be removed by centering. The resulting centered form shows that verifier rewards specify not only which responses should be preferred, but also the relative likelihood changes an improved policy should assign to them.
    \item We propose RSPO, a simple stop-gradient objective for diffusion language models. Instead of using rewards only as fixed advantages, RSPO compares the current centered ELBO-based relative score with the reward-implied target and updates according to the remaining gap. We further provide a local first-order analysis showing that the RSPO update admits a finite centered-score target and is first-order equivalent to a matched quadratic target objective.
    \item We validate RSPO on mathematical reasoning and planning benchmarks. Across these settings, RSPO improves diffusion language models over existing dLLM RL baselines, supporting our key observations and demonstrating the superior performance of the proposed algorithm.
\end{itemize}

\section{Preliminaries}
\label{sec:preliminaries}

\paragraph{Masked Diffusion Language Models.}
Masked diffusion models (MDMs)~\citep{sahoo2024simple} generate sequences by reversing a gradual masking process.
Given a prompt $\bm{q}$ and a clean response $\bm{y}=(y^1,\ldots,y^L)$, the forward process corrupts $\bm{y}$ into $\bm{z}_t=(z_t^1,\ldots,z_t^L)$ at timestep $t\in[0,1]$, while keeping $\bm{q}$ fixed.
Each token is independently replaced by the mask token $[\textsc{mask}]$, whose one-hot representation is denoted by $\bm{m}$: \(q_{t|0}(z_t^i \mid y^i)
=
\mathrm{Cat}\left(
z_t^i;\,
\alpha_t y^i + (1-\alpha_t)\bm{m}
\right),\)
where $\mathrm{Cat}$ denotes a categorical distribution and $\alpha_t$ is a decreasing noise schedule with $\alpha_0=1$ and $\alpha_1=0$.
Thus $\bm{z}_0=\bm{y}$ and $\bm{z}_1$ is fully masked; for the linear schedule $\alpha_t=1-t$, each token is masked with probability $t$.
The reverse process is parameterized by a denoising model $\pi_\theta$.
For $0\le s<t\le 1$, the transition from $\bm{z}_t$ to $\bm{z}_s$ keeps unmasked tokens fixed and predicts masked tokens from the corrupted context:
\begin{equation*}
p_\theta(z_s^i \mid \bm{z}_t,\bm{q})
=
\begin{cases}
\mathrm{Cat}(z_s^i;\, z_t^i),
& z_t^i \ne \bm{m}, \\[1mm]
\mathrm{Cat}\left(
z_s^i;\,
\dfrac{(1-\alpha_s)\bm{m}
+(\alpha_s-\alpha_t)\pi_\theta(\cdot \mid \bm{z}_t,\bm{q})}
{1-\alpha_t}
\right),
& z_t^i = \bm{m}.
\end{cases}
\end{equation*}
Generation starts from a fully masked response and iteratively denoises masked positions, often in parallel.
Since MDMs do not use an autoregressive factorization, their likelihood is commonly optimized through an evidence lower bound (ELBO)~\citep{shi2024simplified,ouyour}:
\begin{equation*}
\mathcal{L}_{\mathrm{ELBO}}(\bm{y}\mid\bm{q};\theta)
=
\mathbb{E}_{t\sim\mathcal{U}[0,1],\,
\bm{z}_t\sim q_{t|0}(\cdot\mid\bm{y})}
\left[
\sum_{i=1}^{L}
w(t)\,
\mathbf{1}[z_t^i=\bm{m}]\,
\log \pi_\theta(y^i\mid\bm{z}_t,\bm{q})
\right]
\le
\log \pi_\theta(\bm{y}\mid\bm{q}) .
\end{equation*}
Here the model is trained to recover clean tokens only at masked positions.
The time-dependent weight \(w(t)=\frac{\alpha_t'}{\alpha_t-1}\) is determined by the noise schedule,
which reduces to $w(t)=1/t$ under the linear schedule $\alpha_t=1-t$.

\paragraph{Reinforcement Learning with Verifiable Rewards.}
In reinforcement learning with verifiable rewards (RLVR), a language model is treated as a policy $\pi_\theta$ that generates a response $\bm{y}$ conditioned on a prompt $\bm{q}$. 
A verifier assigns a scalar reward $r(\bm{y},\bm{q})$ to the completed response, and the goal is to improve the policy toward high-reward outputs while controlling its deviation from a reference policy $\pi_{\mathrm{ref}}$. 
The standard reward-maximization objective is \(    J(\theta)
    =
    \mathbb{E}_{\bm{y}\sim \pi_\theta(\cdot\mid \bm{q})}
    \left[
    r(\bm{y},\bm{q})
    \right],\)
whose policy-gradient form is 
\begin{equation*}
\label{eq:standard-policy-gradient}
    \nabla_\theta J(\theta)
    =
    \mathbb{E}_{\bm{y}\sim \pi_\theta(\cdot\mid \bm{q})}
    \left[
    r(\bm{y},\bm{q})\,
    \nabla_\theta \log \pi_\theta(\bm{y}\mid \bm{q})
    \right].
\end{equation*}
For autoregressive language models, the sequence log-likelihood $\log \pi_\theta(\bm{y}\mid\bm{q})$ can be computed by summing token-level log-probabilities. However, this is no longer tractable for dLLMs, whose generation marginalizes over iterative denoising paths rather than following a fixed left-to-right factorization. 
Consequently, recent dLLM RL methods use an ELBO-based score derived from the denoising objective as a practical substitute ~\citep{zhao2025d1,ou2025principled,wang2025revolutionizing,wang2025spg,lin2025boundary}
\(   \log \pi_\theta(\bm{y}\mid\bm{q})
    \approx
    \mathcal{L}_{\mathrm{ELBO}}(\bm{y}\mid\bm{q};\theta).\) 
Here the ELBO score is tractable because it only requires evaluating denoising probabilities at sampled timesteps and masked positions, rather than summing over all possible denoising trajectories.
\paragraph{The Intractability Challenge.}
This substitution changes what the RL objective can reliably measure. The verifier reward remains a direct task-level signal: it tells us whether a completed response is correct or useful. By contrast, the ELBO-based score is only a tractable proxy for the sequence log-likelihood. It is a lower-bound-style estimate, and in practice it is evaluated through sampled timesteps, sampled corruptions, and token-level denoising predictions.
Thus, the reward signal and the model-side score have different reliability: the former can be exact for verifiable tasks, while the latter is approximate and can be noisy.

This distinction matters when the ELBO score is used inside a reward-weighted policy objective. A high reward can increase the weight placed on a sampled response, but the objective does not know
whether the corresponding ELBO-based relative score is still too small, already sufficiently large, or large mainly because of estimation noise.
As a result, reward weighting provides a direction of preference but not a calibrated target for the relative score itself. The goal of this work is to develop an RLVR objective for dLLMs that uses reliable verifier rewards to calibrate ELBO-based relative scores, rather than merely using rewards as fixed weights on approximate likelihood surrogates.

\section{Method}
\label{sec:method}
\subsection{Motivation}

Section~\ref{sec:preliminaries} motivates the following objective-design issue: when dLLM RLVR relies on an ELBO-based likelihood surrogate, the surrogate requires a well-defined optimization target. We derive this target from the standard KL-regularized policy-improvement problem.
In particular, for a fixed prompt $\bm{q}$ and reference policy $\pi_{\mathrm{ref}}$, the solution of the standard KL-regularized policy-improvement problem is
\begin{equation*}
    \pi^\star(\bm y\mid\bm q)
    =
    Z(\bm q)^{-1}\pi_{\mathrm{ref}}(\bm y\mid\bm q)
    \exp\!\left(r(\bm y,\bm q)/\beta\right).
\end{equation*}
where $Z(\bm q)$ is the prompt-dependent normalizer.
Therefore, the ideal current-reference log-ratio is $\Delta^\star(\bm y,\bm q):=r(\bm y,\bm q)/\beta-\log Z(\bm q)$. For a group $\{\bm y_i\}_{i=1}^G$ sampled from the same prompt, let
$r_i=r(\bm y_i,\bm q)$ and
$\Delta_i^\star=\Delta^\star(\bm y_i,\bm q)$. Then define $\bar \Delta^\star $ by the mean of $\Delta_i^\star$ over all $i$'s, then the normalization factor $Z(\bm q)$ eliminates under centering:
\begin{equation}\label{eq:centered-kl-target}
    \Delta_i^\star-\bar\Delta^\star
    =
    \frac{1}{\beta}(r_i-\bar r).
\end{equation}
In practice, we use a group-relative advantage $\Atilde_i$ as the normalized version of the centered reward signal.
The scaling by $1/\beta$ and any group-shared reward normalization are absorbed into $\Atilde_i$, with $\sum_{i=1}^{G}\Atilde_i=0$ for a complete prompt group.\footnote{The analysis only requires the advantages in a complete prompt group to sum to zero; any group-shared reward scaling preserves this property.} Moreover, we introduce a fixed feedback coefficient $\lambda>0$ and use $\tau_i := \Atilde_i/\lambda$ as the reward-implied target for the centered relative score.
The coefficient $\lambda$ controls the scale of the target: larger $\lambda$ makes the target more conservative, while smaller $\lambda$ asks the model to move farther from the reference for the same group-relative advantage.

It remains to define a tractable relative score for dLLMs. Since exact sequence likelihoods are unavailable, RSPO uses current-reference differences of ELBO scores. Let $\mathcal E_i^\theta=\mathcal{L}_{\mathrm{ELBO}}(\bm{y}_i\mid\bm{q};\theta)$ and $\mathcal E_i^{\mathrm{ref}}=\mathcal{L}_{\mathrm{ELBO}}(\bm{y}_i\mid\bm{q};\mathrm{ref})$. For completion length $L_c$,
\begin{equation}
\label{eq:delta-definition-new}
    \delta_i(\theta)
    =
    \frac{\mathcal E_i^\theta-\mathcal E_i^{\mathrm{ref}}}{L_c}.
\end{equation}
Positive $\delta_i(\theta)$ means that the current model assigns a higher ELBO score to response $\bm y_i$ than the reference model.
If an implementation stores cross-entropy or negative ELBO rather than ELBO, the current-reference difference must be reversed.
\begin{remark}[Sign convention]\label{rem:sign-convention-new}
Eq.~\eqref{eq:delta-definition-new} must increase with completion likelihood; otherwise high-reward samples receive the wrong update direction.
\end{remark}

We then center this score within the same prompt group.
For a group $\mathcal B$ of size $G$, define $\textstyle{\dbar_{\mathcal B}(\theta)=N^{-1}\sum_{j\in\mathcal B}\delta_j(\theta)}$ and $\dhat_i(\theta)=\delta_i(\theta)-\sg(\dbar_{\mathcal B}(\theta))$, where $\sg(\cdot)$ denotes detachment.  Then, The centered score $\dhat_i(\theta)$ is RSPO's tractable counterpart of the centered log-ratio in Eq.~\eqref{eq:centered-kl-target}. When the micro-batch advantages sum to zero, the verifier supplies the target $\dhat_i(\theta)\to \Atilde_i/\lambda$, so the update should follow the residual error between the current relative score and this reward-implied target.

\subsection{Relative Score Policy Optimization}
The name \emph{Relative Score Policy Optimization} reflects the two ingredients above.
First, the quantity being controlled is the centered current-reference relative score $\dhat_i(\theta)$.
Second, the objective keeps the policy-optimization form: a differentiable model-side score is multiplied by a scalar feedback coefficient.

To see the difference from standard advantage weighting, consider a loss micro-batch $\mathcal B$ with $N=|\mathcal B|$.
The advantage-weighted (AW) surrogate uses the group-relative advantage as a fixed coefficient, $\ell_{\mathrm{AW}}(\theta;\mathcal B)=-N^{-1}\sum_{i\in\mathcal B}\Atilde_i\,\dhat_i(\theta)$.
This objective always pushes a positive-advantage response upward, regardless of its current centered relative score.
It therefore encodes a preference direction, but not a stopping rule tied to the reward-implied target.

RSPO replaces this fixed coefficient with the remaining calibration gap.
Recall that the target relative score is $\Atilde_i/\lambda$, or equivalently that the scaled residual is $e_i(\theta)=\Atilde_i-\lambda\dhat_i(\theta)$.
RSPO uses a detached version of this residual as the feedback coefficient, $w_i=\Atilde_i-\lambda\,\sg(\dhat_i(\theta))$, and defines
\begin{equation}
\label{eq:rspo-loss}
    \ell_{\mathrm{RSPO}}(\theta;\mathcal B)
    =
    -\frac{1}{N}\sum_{i\in\mathcal B}
    w_i\,\dhat_i(\theta).
\end{equation}
This coefficient adapts to the current relative score.
If $\dhat_i(\theta)<\Atilde_i/\lambda$, then $w_i>0$ and the update increases the score.
If $\dhat_i(\theta)>\Atilde_i/\lambda$, then $w_i<0$ and the update decreases the score.
Near the target, the coefficient becomes small, so the update naturally weakens.
Thus RSPO keeps the familiar policy-gradient-style objective while turning verifier rewards into calibrated feedback toward a finite relative-score target.
The exact first-order gradient of Eq.~\eqref{eq:rspo-loss} is analyzed in Section~\ref{sec:theory}.

\begin{remark}
A direct squared calibration loss between $\dhat_i(\theta)$ and $\Atilde_i/\lambda$ would induce the same local first-order coefficient.
RSPO deliberately keeps a policy-optimization form instead: the calibration gap is detached and used as a scalar coefficient, while differentiation passes through a single current relative-score factor.
\end{remark}

\subsection{Training Procedure}
\label{sec:training-recipe}

Operationally, RSPO can be implemented with the same outer loop as standard group-relative RLVR, as illustrated in Figure~\ref{fig:RSPO}.
For each prompt, we sample a group of completions, evaluate them with a verifier, and convert the resulting rewards into group-relative advantages.
We then estimate the current-reference ELBO difference for each completion, center these relative scores within the loss micro-batch, and minimize the RSPO loss in Eq.~\eqref{eq:rspo-loss}.
When computing the current and reference ELBO scores for the same completion, we use the same diffusion masks for both models to reduce Monte Carlo variance in their difference.
The reference model is kept fixed throughout training, and $\lambda$ is treated as a fixed feedback-scale hyperparameter.
Full score-estimation and algorithmic details are deferred to Appendix~\ref{app:training-details}.

\section{Theory}
\label{sec:theory}

The previous section introduced RSPO as a simple change to the usual group-relative RLVR update: keep the policy-score form, but replace the fixed advantage coefficient with a residual to a reward-implied relative-score target. This section explains what that change means mathematically by analyzing the local update RSPO applies in one backward pass.
The point is not to prove global convergence for dLLMs, and it is not to assume that we can recover the exact sequence likelihood that Section~\ref{sec:preliminaries} identified as intractable. Instead, we fix the sampled micro-batch: completions, verifier rewards, group-relative advantages, reference ELBO scores, diffusion masks, and the feedback scale $\lambda$ are all treated as constants. The batch center is detached, as in Eq.~\eqref{eq:rspo-loss}, and the only differentiated quantity is the current-model score through $\delta_i(\theta)$.
Under this local view, the theory answers three questions that mirror the design choices in Section~\ref{sec:method}: what centering removes, what coefficient multiplies the current score derivative, and what finite target the update is trying to reach.
Estimator-level details and proofs are deferred to Appendices~\ref{app:theory-details} and~\ref{app:proofs}.

\subsection{Relative-Score Feedback and Fixed Point}
\label{sec:gradient-formula}

We first make precise why RSPO is a \emph{relative-score} method.
The ELBO values used by dLLMs are only tractable surrogates for sequence log-likelihoods, and their absolute level can contain shifts shared by many samples in a batch.
Such shared shifts should not determine which completion is preferred within a prompt group.
Detached centering removes them in the forward pass, while preserving the per-sample derivative of the current score: $\sum_{i\in\mathcal B}\dhat_i=0$, and $\nabla_\theta\dhat_i=\nabla_\theta\delta_i(\theta)$ because the center is not differentiated.

\begin{lemma}[Forward centering and zero-sum weights]
\label{lem:zero-sum-weights}
For a fixed sampled micro-batch, $\sum_{i\in\mathcal B}\dhat_i=0$ and $\sum_{i\in\mathcal B}w_i=\sum_{i\in\mathcal B}\Atilde_i$. If $\mathcal B$ is a union of complete prompt groups and $\Atilde_i$ is a zero-sum group-relative advantage, optionally divided by a group-shared reward scale, then the RSPO weights are also zero-sum.
\end{lemma}

The lemma is a useful consistency check.
When the advantages are group-relative, the feedback term does not introduce a new batch-level push.
It continues to compare responses within the same prompt group, which is exactly the comparison structure used to derive the centered target in Section~\ref{sec:method}.

We next look at the actual gradient of the RSPO loss.
Since the residual coefficient is detached, differentiation passes through only one copy of the current relative score.

\begin{proposition}[RSPO relative-score feedback gradient]
\label{prop:rspo-feedback-gradient}
For fixed $\lambda$ and a fixed sampled micro-batch,
\begin{equation}
\label{eq:rspo-feedback-gradient}
    \nabla_\theta\ellRSPO(\theta;\mathcal B)
    =
    -\frac{1}{N}\sum_{i\in\mathcal B}
    (\Atilde_i-\lambda\dhat_i)
    \nabla_\theta\delta_i(\theta).
\end{equation}
The policy-gradient coefficient is the relative-score feedback error, rather than the advantage alone.
\end{proposition}

This formula is the core local interpretation of RSPO.
The usual advantage-weighted update, as used in policy-gradient-style RL objectives such as PPO or GRPO~\citep{schulman2017proximal,shao2024deepseekmath}, multiplies the score derivative by a fixed advantage.
RSPO instead multiplies it by $\Atilde_i-\lambda\dhat_i$.
If the centered score is still below its reward-implied target $\Atilde_i/\lambda$, the coefficient is positive and the update increases the score.
If the score has already passed that target, the coefficient becomes negative and pulls it back.
Near the target, the coefficient is small.
So the update is still a policy-score update, but the coefficient now measures how much target error remains.

\begin{corollary}[Standard advantage-weighted objective at the reference point]
\label{cor:centered-advantage-limit}
If the coupled current and reference ELBO scores match for every sampled completion in the micro-batch, then $\dhat_i=0$ for all $i\in\mathcal B$ and $\nabla_\theta\ellRSPO(\theta;\mathcal B)=\nabla_\theta\ellAW(\theta;\mathcal B)$. If additionally $\sum_{i\in\mathcal B}\Atilde_i=0$, then centering does not change the forward value of this standard surrogate.
\end{corollary}

This corollary explains why RSPO should be viewed as a modification of advantage-weighted RLVR, not a replacement for the policy-improvement direction.
At the reference point, before the current model has moved away from the reference on the sampled completions, RSPO gives the same first-order update as the standard advantage-weighted surrogate.
The distinction appears after the relative scores start moving: advantage weighting keeps applying the same coefficient, whereas RSPO asks whether the current score is still below, near, or above its target.

\begin{proposition}[Fixed point under zero-sum advantages]
\label{prop:fixed-point}
Assume $\lambda>0$ and $\sum_{i\in\mathcal B}\Atilde_i=0$. If
\begin{equation}
\label{eq:fixed-point-condition}
    \dhat_i = \Atilde_i/\lambda,
    \qquad i\in\mathcal B,
\end{equation}
then every sample weight is zero and $\nabla_\theta\ellRSPO(\theta;\mathcal B)=0$. Conversely, if the Jacobian of $\delta(\theta)$ has full row rank on the micro-batch, every first-order stationary point satisfies Eq.~\eqref{eq:fixed-point-condition}.
\end{proposition}

The fixed point is the formal version of the target-tracking story.
A reward advantage is not treated as a coefficient to apply indefinitely.
It specifies a finite centered current-reference score, with $\lambda$ setting the scale of that target.
The full-row-rank assumption is only needed for the converse direction; even without it, Eq.~\eqref{eq:fixed-point-condition} is the condition under which all RSPO feedback weights vanish.

\subsection{First-Order Equivalence of RSPO}
\label{sec:first-order-equivalence-rspo}

The preceding results describe the coefficient used by RSPO.
There is still a natural question: should RSPO be understood as a calibration objective, or as a policy-optimization update?
The answer is that, locally, it is both.
It tracks the same target as a simple quadratic calibration loss, while keeping the implementation in the familiar form of a detached coefficient times a differentiable policy score.

Let $\delta$ and $\dhat$ denote the micro-batch vectors, and write $\innerG{a}{b}=N^{-1}\sum_{i\in\mathcal B}a_i b_i$ and $\normG{a}^2=\innerG{a}{a}$.
Consider the matched quadratic objective
\begin{equation}
\label{eq:quadratic-rspo}
    \ell_{\mathrm{quad},\lambda}(\theta;\mathcal B)
    =
    -\innerG{\Atilde}{\delta}
    +
    \frac{\lambda}{2}\normG{\dhat}^2.
\end{equation}
When the advantages are zero-sum, completing the square shows that Eq.~\eqref{eq:quadratic-rspo} has the same target $\dhat_i=\Atilde_i/\lambda$ as Proposition~\ref{prop:fixed-point}, up to terms whose derivative vanishes under detached centering; see Appendix~\ref{app:coefficient-convention}.
This comparison separates the target being tracked from the computational graph used to track it.

\begin{theorem}[First-order equivalence to a matched quadratic objective]
\label{thm:first-order-equivalence-rspo}
For fixed $\lambda$ and a fixed sampled micro-batch,
\begin{equation}
\label{eq:rspo-quadratic-gradient-equivalence}
    \nabla_\theta\ellRSPO(\theta;\mathcal B)
    =
    \nabla_\theta\ell_{\mathrm{quad},\lambda}(\theta;\mathcal B).
\end{equation}
Thus RSPO follows the same first-order update vector as Eq.~\eqref{eq:quadratic-rspo}, while detaching one copy of $\dhat$ in the feedback term.
\end{theorem}

The theorem clarifies the role of the stop-gradient construction.
The finite target can be written as a quadratic calibration target, but RSPO implements the corresponding first-order direction as a policy-score update using the quantities already computed during training: a centered current-reference ELBO score and a detached residual coefficient.
The equivalence is only first-order.
The scalar objectives differ for higher-order differentiation, and this section does not claim a general variance reduction or global convergence guarantee.
Instead, it gives the local mechanism that the experiments probe next: whether residual feedback, reference subtraction, and centering improve dLLM RLVR when the likelihood signal is an ELBO-based surrogate.

\section{Experiments}
\label{sec:experiments}
The experiments are designed to test the two claims suggested by the method and theory.
First, if verifier rewards can serve as targets for centered relative scores, then RSPO should be especially useful in settings where the reward is reliable but sparse: the verifier can judge a completed answer, but it does not provide token-level or denoising-step supervision.
Second, if the gains come from the relative-score feedback mechanism, then removing feedback, reference subtraction, or centering should change the optimization behavior in predictable ways.
We therefore evaluate RSPO on both planning and mathematical reasoning benchmarks, and then use ablations to isolate the roles of feedback, current-reference scoring, and centering.

\subsection{Experimental Setup}
\paragraph{Models and Datasets.}
We follow the experimental protocols of d1~\citep{zhao2025d1} and wd1~\citep{tang2025wd1}. We use LLaDA-8B-Instruct~\citep{nielarge} as the base diffusion language models. Our evaluation covers two categories of reasoning tasks: mathematical reasoning, including GSM8K~\citep{cobbe2021training} and MATH500~\citep{lightman2023let}, and planning, including Countdown~\citep{tinyzero} and Sudoku~\citep{cordero_sudoku_generator}. We follow prior work~\citep{zhao2025d1,tang2025wd1} for train-test splits, reward functions, and evaluation protocols, except for Sudoku. For Sudoku, to avoid train-test leakage, we split the data by puzzle solutions so that the test set contains unseen answers, preventing models from solving test instances through memorization.

\paragraph{Baselines.}
We compare RSPO with representative RL fine-tuning methods for diffusion language models, including d1/Diffu-GRPO~\citep{zhao2025d1}, VRPO on LLaDA-1.5~\citep{zhu2025llada}, wd1~\citep{tang2025wd1}, SAPO~\citep{xie2025step}, and TraceRL~\citep{wang2025revolutionizing}. These baselines cover both GRPO-style adaptations and recent dLLM-specific policy optimization methods, providing a broad comparison across mathematical and planning benchmarks.

\paragraph{Training Details.}
For RSPO training, we apply Low-Rank Adaptation (LoRA) with rank $r=128$ and scaling factor $\alpha=64$, and set \(\lambda=0.01\). All experiments are conducted in the zero-shot setting. For both RL rollouts and evaluation, we use the semi-autoregressive confidence-based decoding strategy following LLaDA, d1, and wd1. We apply the same generation setup as d1: the denoising timestep is set to half of the total sequence length, and the sequence is divided into blocks of 32 tokens. At each diffusion step, we unmask the two tokens with the highest confidence, measured by the probability of the sampled token, within the current incomplete block. During RL rollouts, we use a generation length of 256 and a sampling temperature of 0.9 across all benchmarks, except for Sudoku, where the temperature is set to 0.3 following d1. During evaluation, the sampling temperature is set to 0.0. We evaluate checkpoints every 100 training steps and report results from the checkpoint with the highest average test accuracy across generation lengths of 256 and 512.

\subsection{Main Results}
Table~\ref{tab:main_results} reports final accuracy on planning and mathematical reasoning benchmarks.
RSPO gives its clearest gains on sparse-reward planning tasks: on Sudoku, it reaches 92.1/90.8 accuracy at generation lengths 256/512, improving over the strongest non-RSPO result by 66.5/65.4 points; on Countdown, it reaches 78.83/73.83, improving by 26.83/17.53 points.
These gains are consistent across generation budgets and align with the motivation of RSPO: reliable response-level rewards can serve as calibrated targets for noisy ELBO-based current-reference scores.
On mathematical reasoning, the pattern is more mixed but still favorable.
RSPO obtains the best length-256 results on both GSM8K and MATH500, and the best length-512 result on MATH500.
On GSM8K at length 512, RSPO is below the strongest baseline, indicating that the benefit is task-dependent in high-performing mathematical-reasoning regimes.
\emph{Overall, RSPO is strongest on planning benchmarks while remaining competitive on standard mathematical reasoning tasks.}

\begin{figure*}[t]
    \centering
    \includegraphics[width=0.45\linewidth]{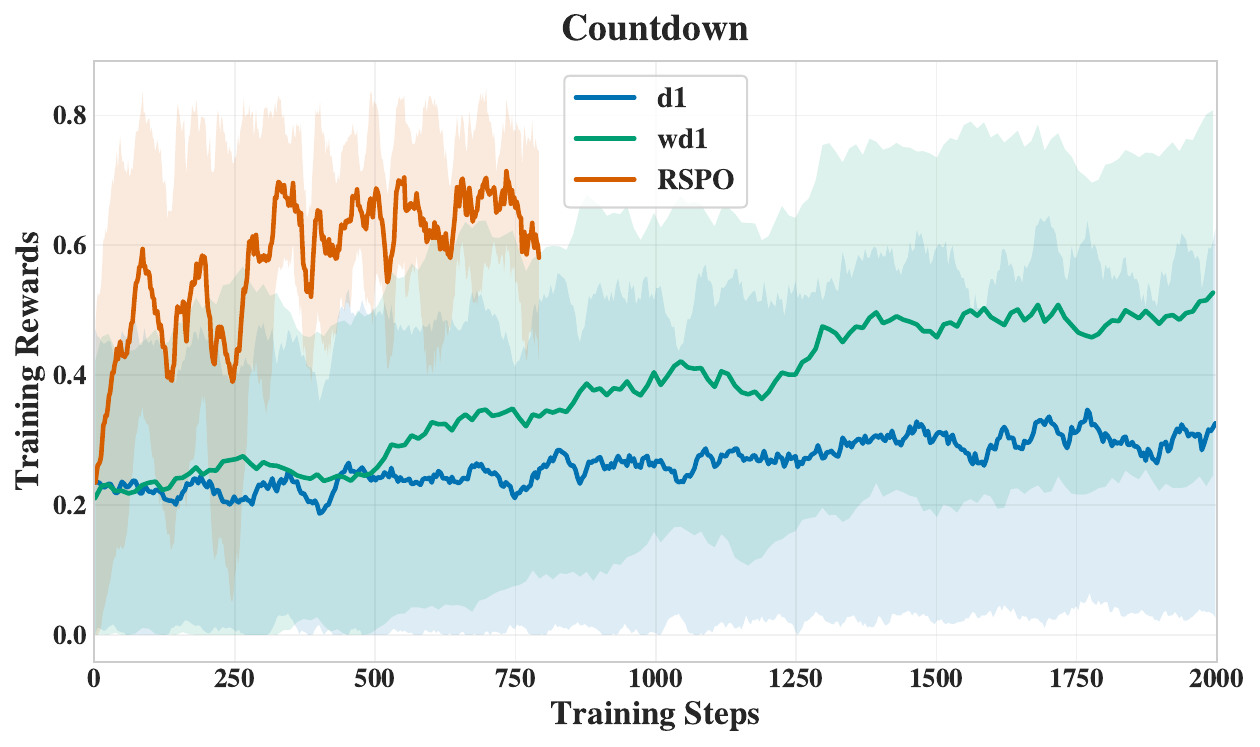}
    \hspace{.5cm}
    \includegraphics[width=0.45\linewidth]{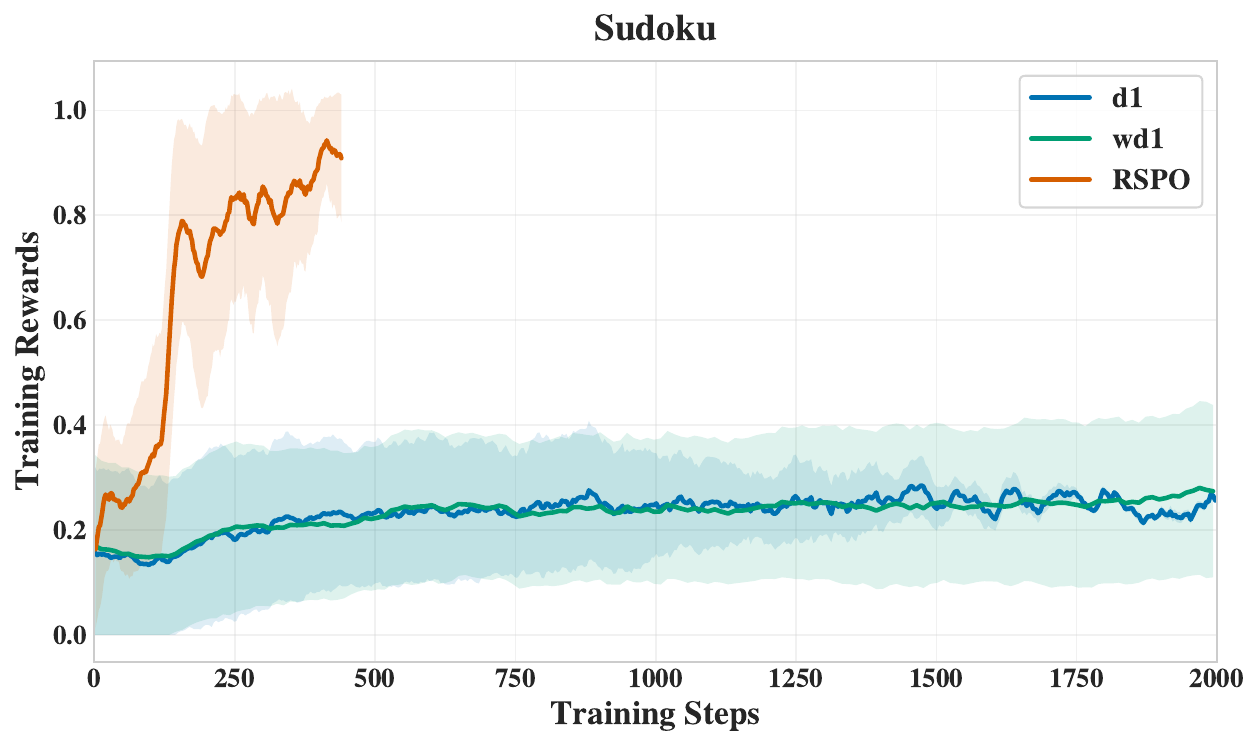}
    \vspace{-5pt}
    \caption{Training reward dynamics of RSPO and baselines on planning benchmarks. Shaded regions indicate variation across runs.}
    \label{fig:reward-dynamics}
    \vspace{-10pt}
\end{figure*}

Figure~\ref{fig:reward-dynamics} reports training reward trajectories on the planning benchmarks.
On both Countdown and Sudoku, RSPO improves verifier rewards faster than d1 and wd1 and reaches higher reward levels within the plotted training window.
This trajectory evidence suggests that RSPO changes the optimization path during training, rather than merely selecting a better final checkpoint.
It also matches the local mechanism in Section~\ref{sec:theory}: as a response approaches its reward-implied relative-score target, RSPO reduces or reverses the update coefficient instead of repeatedly applying the same fixed advantage.
Additional reasoning reward trajectories are provided in Figure~\ref{fig:reward-dynamics-math} in Appendix~\ref{app:experimental-details}.

\begin{table}[t]
\centering
\caption{Benchmark accuracy of RSPO and dLLM RL baselines across planning and mathematical reasoning tasks. Columns 256 and 512 denote generation length; best results are in \textbf{bold}.}
\label{tab:main_results}
\small
\setlength{\tabcolsep}{5.5pt}
\renewcommand{\arraystretch}{1.08}
\begin{adjustbox}{max width=\textwidth}
\begin{tabular}{lcccccccc}
\toprule
\multirow{2}{*}{Methods} 
& \multicolumn{2}{c}{Sudoku}
& \multicolumn{2}{c}{Countdown}
& \multicolumn{2}{c}{GSM8K}
& \multicolumn{2}{c}{Math500} \\
\cmidrule(lr){2-3}
\cmidrule(lr){4-5}
\cmidrule(lr){6-7}
\cmidrule(lr){8-9}
& 256 & 512 & 256 & 512 & 256 & 512 & 256 & 512 \\
\midrule
LLaDA-8B-Instruct~\citep{nielarge}
& 16.2 & 6.0 & 19.5 & 16.0 & 76.7 & 78.2 & 32.4 & 36.2 \\

+ d1~\citep{zhao2025d1}
& 24.1 & 15.9 & 31.3 & 37.1 & 78.1 & 81.2 & 34.1 & 39.0 \\

+ VRPO~\citep{zhu2025llada}
& 12.8 & 9.6 & 22.3 & 18.0 & 80.1 & 81.5 & 35.6 & 34.8 \\

+ wd1~\citep{tang2025wd1}
& 22.0 & 24.6 & 51.2 & 46.1 & 80.8 & 82.3 & 34.4 & 39.0 \\

+ SAPO~\citep{xie2025step}
& 20.3 & 16.1 & 52.0 & 56.3 & 80.6 & 82.1 & 33.8 & 38.4 \\

+ TraceRL~\citep{wang2025revolutionizing}
& 25.6 & 25.4 & 50.4 & 52.6 & 81.3 & \best{82.4} & 35.6 & 39.1 \\


\rowcolor{gray!18}
+ RSPO (ours)
& \best{92.1}
& \best{90.8}
& \best{78.83}
& \best{73.83}
& \best{82.63}
& 80.55
& \best{42.07}
& \best{40.25} \\
\bottomrule
\end{tabular}
\end{adjustbox}
\vspace{-8pt}
\end{table}

\subsection{Ablation Studies}
\label{sec:ablations}
The ablations focus on Sudoku\_new~\citep{wang2025spg}, a sparse correctness-reward planning task, to isolate the three RSPO components introduced in Sections~\ref{sec:method} and~\ref{sec:theory}: residual feedback through $\lambda$, current-reference subtraction in $\delta_i$, and centering of the relative scores.

\begin{figure}[t]
    \centering
    \begin{subfigure}[t]{0.45\linewidth}
        \centering
        \includegraphics[width=\linewidth]{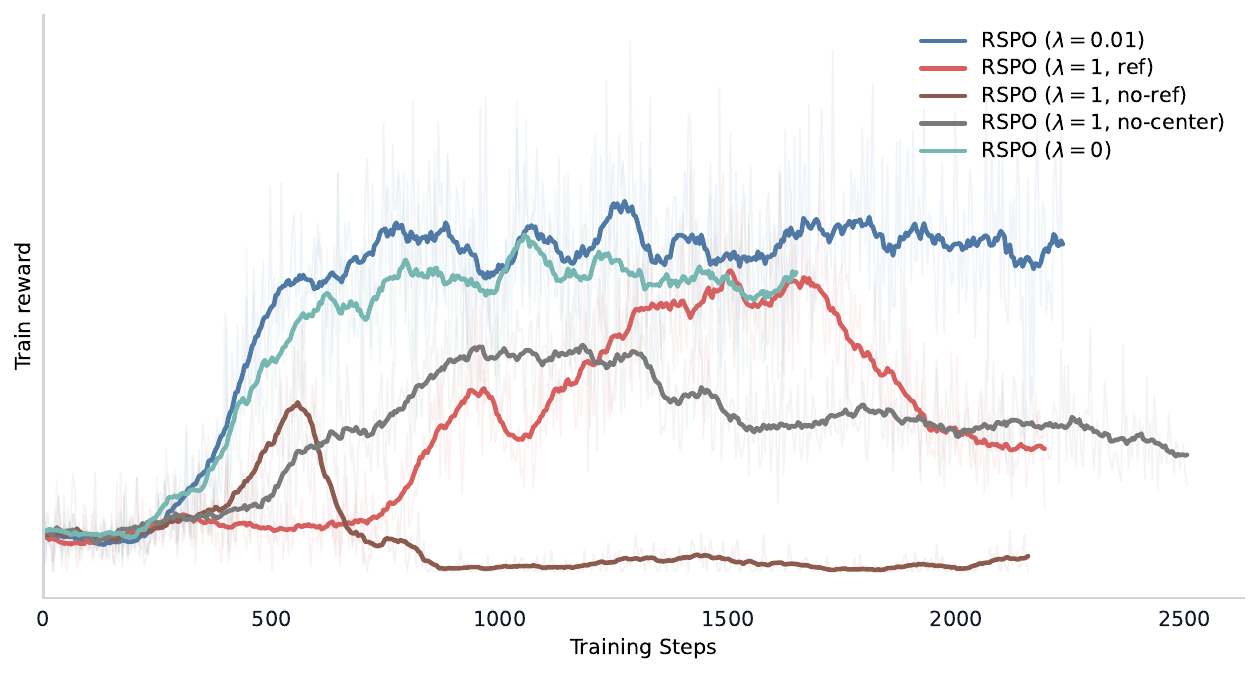}
        \caption{Training reward across feedback variants}
        \label{fig:ablation-sudoku-reward}
    \end{subfigure}
    \hspace{.5cm}
    \begin{subfigure}[t]{0.45\linewidth}
        \centering
\includegraphics[width=\linewidth]{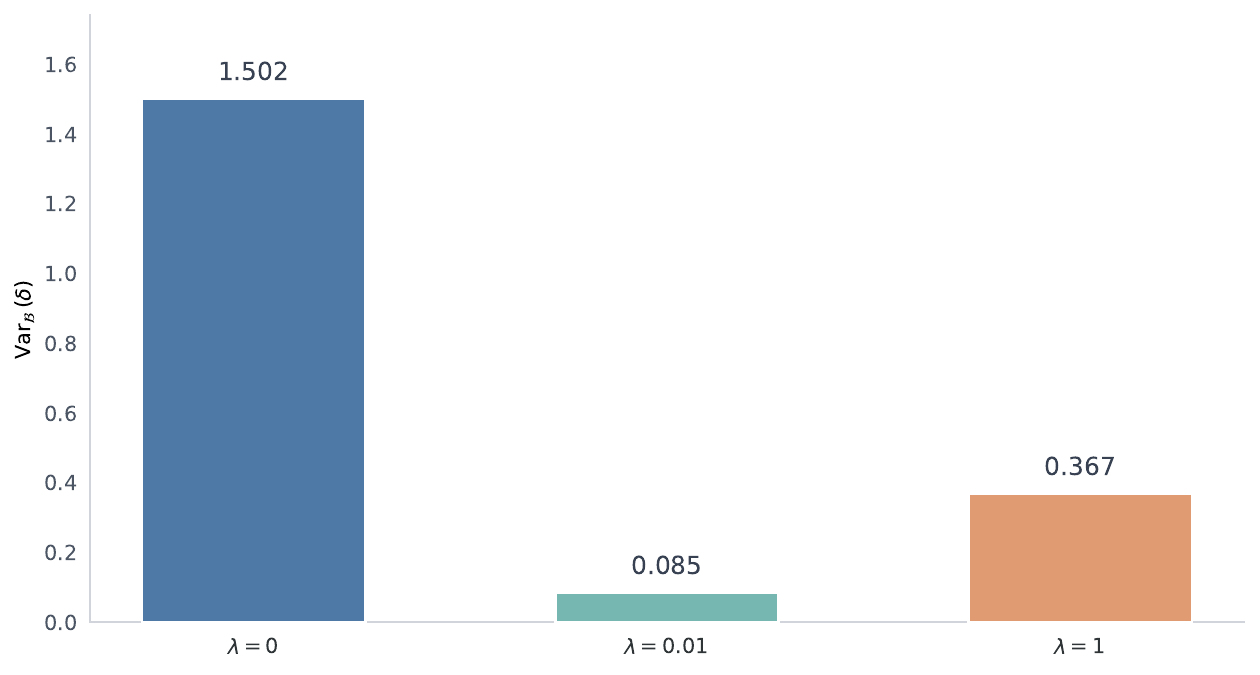}
        \caption{Variance of centered relative scores $\VarG(\delta)$}
        \label{fig:ablation-sudoku-var}
    \end{subfigure}
      \vspace{-5pt}
    \caption{Sudoku ablations of RSPO feedback components. The two panels compare reward dynamics and centered relative-score variance under different feedback settings.}
    \label{fig:sudoku-ablation}
    \vspace{-10pt}
\end{figure}

\paragraph{Relative-Score Feedback.}
We first ablate the feedback coefficient.
Setting $\lambda=0$ removes the residual feedback term and reduces the update to the standard advantage-weighted form over the centered current-reference score.
Figure~\ref{fig:ablation-sudoku-reward} shows that this variant improves early but converges to a lower reward level, supporting Section~\ref{sec:method}: fixed advantages rank samples but do not define finite relative-score targets.
RSPO with $\lambda=0.01$ gives the best reward dynamics, whereas stronger feedback at $\lambda=1$ is less effective and removing reference-model subtraction at $\lambda=1$ severely degrades reward.
\emph{Thus RSPO works best when feedback calibrates current-reference scores without letting noisy ELBO estimates dominate reward-driven improvement.}

\paragraph{Relative-Score Variance Diagnostics.}
We next examine the spread of the centered current-reference scores.
Figure~\ref{fig:ablation-sudoku-var} reports $\VarG(\delta)$, the batch variance of the ELBO-based relative scores after centering.
Because centering removes shared score offsets, this quantity measures relative spread within the response group rather than a batch-level shift.
Without feedback ($\lambda=0$), $\VarG(\delta)$ is large; at $\lambda=1$, the spread is reduced but rewards are worse.
The moderate setting $\lambda=0.01$ gives both the strongest reward dynamics and the smallest measured $\VarG(\delta)$, consistent with Proposition~\ref{prop:rspo-feedback-gradient}.
\emph{Thus the diagnostic supports feedback calibration, but remains only a proxy related to Appendix~\ref{sec:kl-interpretation} and is not used to adapt $\lambda$.}

\paragraph{Batch-Mean Score-Offset Diagnostics.}
Finally, we check the centering operation itself.
RSPO applies feedback to $\dhat_i$, not to the raw ELBO difference $\delta_i$, so shared ELBO-score shifts do not affect within-group preferences.
Table~\ref{tab:score-offset-ablation} reports that the mean absolute batch-mean offset stays near zero with centering, including $1.32{\times}10^{-9}$ at $\lambda=0.01$ and $9.91{\times}10^{-9}$ at $\lambda=1$, while removing centering raises it to $1.14{\times}10^{-1}$ with much larger variation.
\emph{This confirms the centering condition in Lemma~\ref{lem:zero-sum-weights}: RSPO feedback should act on relative likelihood changes within the response group, rather than on a shared ELBO-score shift.}

\begin{table}[t]
\centering
\caption{Batch-mean relative-score offset in Sudoku ablations. The table checks whether centering removes shared ELBO-score shifts across feedback and reference settings.}
\label{tab:score-offset-ablation}
\small
\setlength{\tabcolsep}{4.5pt}
\renewcommand{\arraystretch}{1.08}
\begin{adjustbox}{max width=\textwidth}
\begin{tabular}{cccccc}
\toprule
\(\lambda\) & Centering & Reference & Mean \( |\mathrm{offset}| \) & Std. & Range \\
\midrule
0 & $\checkmark$ & $\checkmark$ & $1.80{\times}10^{-8}$ & $2.40{\times}10^{-8}$ & $[-1.14,\,1.23]{\times}10^{-7}$ \\
0.01 & $\checkmark$ & $\checkmark$ & $1.32{\times}10^{-9}$ & $2.41{\times}10^{-9}$ & $[-1.68,\,1.49]{\times}10^{-8}$ \\
1 & $\checkmark$ & $\checkmark$ & $9.91{\times}10^{-9}$ & $1.28{\times}10^{-8}$ & $[-5.59,\,5.59]{\times}10^{-8}$ \\
\midrule
1 & $\checkmark$ & -- & $3.94{\times}10^{-8}$ & $4.94{\times}10^{-8}$ & $[-1.49,\,1.49]{\times}10^{-7}$ \\
1 & -- & $\checkmark$ & $1.14{\times}10^{-1}$ & $1.42{\times}10^{-1}$ & $[-4.11,\,5.08]{\times}10^{-1}$ \\
\bottomrule
\end{tabular}
\end{adjustbox}
\vspace{-10pt}
\end{table}

\section{Conclusion and Limitations}
\label{sec:discussion}

This paper introduced RSPO, a relative-score objective for RLVR post-training of diffusion language models. The central idea is to use group-relative verifier rewards not only to rank sampled completions, but also to define where their centered current-reference scores should move. RSPO implements this idea by converting the gap between the reward-implied target and the current centered ELBO score into a detached feedback coefficient for a policy-optimization update. This gives a direct way to use noisy dLLM likelihood surrogates without treating reward advantages as indefinitely applied update weights. Empirically, RSPO shows its clearest gains on sparse-reward planning tasks and remains competitive on mathematical reasoning benchmarks, with ablations supporting the roles of residual feedback, reference subtraction, and centering.

We also note several limitations that point to useful future directions. The current configurations of RSPO keep RSPO simple and close to standard group-relative RLVR, but they leave open how to adapt $\lambda$ automatically or vary the reference policy during training. The theory in Section~\ref{sec:theory} is local and first-order: it explains the implemented update by matching RSPO to a quadratic target objective at the gradient level, but does not establish global convergence or higher-order equivalence. It is interesting and important to develop a global theory for dLLM RL with ELBO-based score estimates. Finally, the score-spread diagnostic $\VarG(\delta)$ is only a local proxy rather than an exact KL estimator, and implementations must respect the ELBO sign convention in Remark~\ref{rem:sign-convention-new}; more precise trust-region diagnostics and robust score-estimation procedures are promising directions for future work.
\bibliographystyle{plainnat}
\bibliography{nips}

\newpage
\appendix

\begingroup
\setcounter{tocdepth}{2}
\tableofcontents
\endgroup

\section{Notation Summary}
\label{app:notation}

\begin{table}[h]
\centering
\caption{Notation used in the RSPO formulation.}
\label{tab:notation}
\begin{adjustbox}{max width=\textwidth}
\begin{tabular}{ll}
\toprule
Symbol & Meaning \\
\midrule
$q$ & Prompt \\
$y$ & Generic completion used in policy-improvement discussion \\
$o_i$ & The $i$-th sampled completion in a group \\
$G$ & Group size \\
$K$ & Number of ELBO Monte Carlo mask samples per completion \\
$\mathcal B$ & Current loss micro-batch \\
$N$ & Micro-batch size, $|\mathcal B|$ \\
$\pitheta$ & Current model or policy parameterized by $\theta$ \\
$\piref$ & Frozen reference model or policy \\
$\pi^\star$ & Ideal KL-regularized improved policy in the motivating calculation \\
$\beta$ & KL temperature in the motivating policy-improvement calculation \\
$Z(q)$ & Prompt-level normalizer in the motivating policy-improvement calculation \\
$r_i$ & Reward of completion $o_i$ \\
$\bar r$ & Group mean reward, $G^{-1}\sum_i r_i$ \\
$\sigma_r$ & Group reward standard deviation for the normalized advantage variant \\
$s_q$ & Prompt-level scale used by the normalized advantage variant \\
$\Atilde_i$ & Zero-sum group-relative advantage passed to the loss, optionally divided by $\sigma_r+10^{-4}$ \\
$L_c$ & Number of completion tokens used to normalize score differences \\
$\widehat{\mathcal E}_\theta$ & Likelihood-oriented ELBO or score estimate for the current model \\
$\widehat{\mathcal E}_{\mathrm{ref}}$ & Matched reference-model ELBO or score estimate \\
$M_{i,k}$ & The $k$-th sampled diffusion mask set for completion $o_i$ \\
$\delta_i$ & Per-token ELBO log-ratio surrogate, $(\widehat{\mathcal E}_{\theta}(o_i;q)-\widehat{\mathcal E}_{\mathrm{ref}}(o_i;q))/L_c$ \\
$\dbar_{\mathcal B}$ & Current micro-batch mean of $\delta$ \\
$\dhat_i$ & Centered relative-score surrogate, $\delta_i-\sg(\dbar_{\mathcal B})$ \\
$\lambda$ & Fixed feedback coefficient; the main RSPO experiments set $\lambda=0.01$ unless otherwise stated \\
$\sg(\cdot)$ & Detachment operator \\
$e_i$ & Remaining calibration gap, $\Atilde_i-\lambda\dhat_i$ \\
$w_i$ & RSPO weight, $\Atilde_i-\lambda\sg(\dhat_i)$ \\
$\ellAW$ & Standard group-relative advantage-weighted surrogate \\
$\ellRSPO$ & RSPO objective \\
$\ell_{\mathrm{quad},\lambda}$ & Matched quadratic objective with RSPO's relative-score target \\
$\innerG{a}{b}$ & Micro-batch inner product, $N^{-1}\sum_{i\in\mathcal B}a_i b_i$ \\
$\normG{a}$ & Micro-batch norm, with $\normG{a}^2=\innerG{a}{a}$ \\
$\VarG(\delta)$ & Within-micro-batch variance of the relative-score surrogate \\
\bottomrule
\end{tabular}
\end{adjustbox}
\end{table}

\section{Training Procedure and Score Estimation}
\label{app:training-details}

This appendix specifies the score-estimation and training details used by RSPO. The main text abstracts these details into an ELBO-based current-reference relative score $\delta_i$ for each completion, micro-batch centering, and the relative-score feedback loss in Eq.~\eqref{eq:rspo-loss}.

\subsection{Masked diffusion ELBO estimator}
\label{sec:elbo-estimator}

For a prompt-completion pair $(q,o_i)$, let $L_c$ be the number of completion tokens. We estimate a likelihood-oriented sequence score by sampling $K$ masks and averaging denoising log-probabilities on completion tokens:
\begin{equation}
\label{eq:masked-score}
    \widehat{\mathcal E}_\theta(o_i;q)
    =
    \frac{1}{K}\sum_{k=1}^{K}
    \frac{L_c}{|M_{i,k}\cap o_i|}
    \sum_{t\in M_{i,k}\cap o_i}
    \log p_\theta(x_t \mid x_{\setminus M_{i,k}}, q, M_{i,k}).
\end{equation}
Here $M_{i,k}$ is the $k$-th sampled mask set for completion $o_i$, $x_t$ is the token at position $t$, $x_{\setminus M_{i,k}}$ denotes the sequence context outside the masked positions, and $p_\theta$ is the current model's denoising distribution. Masks with no completion token are resampled or omitted from the Monte Carlo average. For each completion, the same mask set is reused for the current model and the reference model, giving the coupled per-token score difference
\begin{equation*}
    \delta_i
    =
    \frac{\widehat{\mathcal E}_\theta(o_i;q)
    -
    \widehat{\mathcal E}_{\mathrm{ref}}(o_i;q)}
    {L_c}.
\end{equation*}
The factor $L_c/|M_{i,k}\cap o_i|$ makes $\widehat{\mathcal E}_\theta$ a sequence-level completion score, so the division by $L_c$ in $\delta_i$ gives the per-token score difference used in Eq.~\eqref{eq:delta-definition-new}. If an implementation stores a negative ELBO or cross-entropy instead, the sign must follow Remark~\ref{rem:sign-convention-new}.

\subsection{Algorithm}
\label{sec:algorithm}

\begin{algorithm}[h]
\caption{RSPO for diffusion language model reinforcement learning}
\label{alg:rspo}
\begin{algorithmic}[1]
\REQUIRE  Current dLLM $\pi_\theta$, fixed reference dLLM $\pi_{\mathrm{ref}}$, reward function $R$, group size $G$, ELBO samples $K$, feedback coefficient $\lambda>0$.
\FOR{each training step}
    \STATE Sample prompts and generate $G$ completions $\{o_i\}_{i=1}^{G}$ per prompt from $\pi_\theta$.
    \STATE Compute rewards $r_i=R(q,o_i)$ and group-relative advantages within each prompt group.
    \STATE Optionally normalize the advantages within the group to obtain $\Atilde_i$.
    \STATE Estimate $\widehat{\mathcal E}_\theta(o_i;q)$ and $\widehat{\mathcal E}_{\mathrm{ref}}(o_i;q)$ with the shared masks in Eq.~\eqref{eq:masked-score}.
    \STATE Form $\delta_i=(\widehat{\mathcal E}_{\theta}(o_i;q)-\widehat{\mathcal E}_{\mathrm{ref}}(o_i;q))/L_c$ and center it within the loss micro-batch: $\dhat_i=\delta_i-\sg(\dbar_{\mathcal B})$.
    \STATE Set the relative-score feedback weight $w_i=\Atilde_i-\lambda\sg(\dhat_i)$ and descend $\ellRSPO(\theta;\mathcal B)=-N^{-1}\sum_{i\in\mathcal B}w_i\dhat_i$.
    \STATE Record $\VarG(\delta)$ as a stability diagnostic.
\ENDFOR
\end{algorithmic}
\end{algorithm}

\section{Related Work}
\label{app:related-work}

\paragraph{Diffusion Language Models.}
Diffusion models were originally developed as a powerful generative framework for continuous data, especially high-fidelity image generation~\citep{songscore,ho2020denoising,dhariwal2021diffusion} and video generation~\citep{wan2025wan,seedance2026seedance,hacohen2024ltx,hacohen2026ltx,zheng2025open}. 
More recently, this paradigm has been extended to discrete text generation, giving rise to diffusion language models that generate token sequences through iterative denoising~\citep{austin2021structured,campbell2022continuous,loudiscrete,sahoo2024simple,shi2024simplified}.
Unlike autoregressive models, dLLMs are not tied to a fixed left-to-right generation order, enabling bidirectional context modeling, flexible-order refinement, and parallel decoding. 
Recent large-scale dLLMs, such as LLaDA~\citep{nielarge,bie2025llada2}, Dream~\citep{ye2025dream}, multimodal diffusion language models~\citep{yang2025mmada,you2025llada,you2026llada,ai2026llada2}, and block-diffusion variants~\citep{arriolablock,cheng2025sdar}, have demonstrated competitive generation quality and promising efficiency compared with autoregressive models. 
These advances motivate post-training methods that can further improve dLLMs on reasoning, alignment, and task-specific objectives.

\paragraph{Reinforcement Learning for Diffusion Language Models.}
RL for dLLMs is challenging because PPO/GRPO-style objectives rely on policy log-probabilities or importance ratios that are tractable for autoregressive models but unavailable for diffusion generation. 
Existing methods mainly differ in how they approximate or avoid these ratios. 
diffu-GRPO~\citep{zhao2025d1} uses one-step mean-field log-probability estimates to adapt GRPO to masked dLLMs, while wd1~\citep{tang2025wd1} removes explicit importance ratios by reformulating policy optimization as a weighted log-likelihood objective. 
ELBO-based methods approximate likelihoods or log-ratios with denoising objectives, including VRPO in LLaDA~1.5~\citep{zhu2025llada}, sequence-level ELBO policy optimization~\citep{ou2025principled}, sandwiched evidence-bound policy gradients~\citep{wang2025spg}, and memory-efficient large Monte Carlo (large-MC) ELBO optimization~\citep{lin2025boundary}. 
Other works exploit diffusion trajectories or step structure, such as TraceRL~\citep{wang2025revolutionizing} and d-TreeRPO~\citep{pan2025d}, or stabilize noisy ratio-based training with clipping and self-normalization~\citep{zhong2026stabilizing}. 
Our work is complementary: instead of changing the likelihood surrogate, trajectory estimator, or clipping rule, we derive the centered log-ratio target implied by KL-regularized group-relative policy improvement and optimize noisy ELBO log-ratios toward this target with relative-score feedback.

\section{Additional Theory Details}
\label{app:theory-details}

\subsection{Coefficient convention for the matched quadratic objective}
\label{app:coefficient-convention}

Expanding the RSPO loss makes the coefficient convention explicit:
\begin{equation}
\label{eq:rspo-expansion}
    \ellRSPO(\theta;\mathcal B)
    =
    -\innerG{\Atilde}{\dhat}
    +
    \lambda\innerG{\sg(\dhat)}{\dhat}.
\end{equation}
The forward value of the detached feedback term in Eq.~\eqref{eq:rspo-expansion} is $\lambda\normG{\dhat}^2$, but its first-order gradient is the gradient of $\frac{\lambda}{2}\normG{\dhat}^2$. If one removes the detachment operator from the written scalar expression $\lambda\innerG{\sg(\dhat)}{\dhat}$ without changing the coefficient, the penalty gradient doubles. Therefore, comparisons with a fully differentiable matched quadratic objective use the first-order matched coefficient, i.e., $\frac{\lambda}{2}\normG{\dhat}^2$.

Let $c_{\mathcal B}=\sg(N^{-1}\sum_{j\in\mathcal B}\delta_j)$ be the detached micro-batch center and let $\mathbf 1$ be the all-ones vector, so $\dhat=\delta-c_{\mathcal B}\mathbf 1$. Completing the square in Eq.~\eqref{eq:quadratic-rspo} gives the local relative-score target form
\begin{equation}
\label{eq:quadratic-relative-score-target}
    \ell_{\mathrm{quad},\lambda}(\theta;\mathcal B)
    =
    \frac{\lambda}{2}
    \normG{\dhat-\frac{1}{\lambda}\Atilde}^2
    -
    \frac{1}{2\lambda}\normG{\Atilde}^2
    -
    c_{\mathcal B}\innerG{\Atilde}{\mathbf 1}.
\end{equation}
When the micro-batch advantages sum to zero, the final term vanishes and the visible target is $\dhat=\Atilde/\lambda$. If the advantages are not zero-sum, the final term is not a forward-value constant, but it has zero derivative in the local backward pass because $c_{\mathcal B}$ is detached.

Theorem~\ref{thm:first-order-equivalence-rspo} is first-order only. RSPO is not the same scalar objective as Eq.~\eqref{eq:quadratic-rspo} for higher-order differentiation, and the theorem does not by itself establish lower gradient variance. The precise claim is that the update uses the relative-score feedback coefficient $(\Atilde_i-\lambda\dhat_i)$ while preventing one copy of the noisy ELBO difference from contributing a differentiable path in the backward graph. Variance reduction is therefore an empirical property to be evaluated in experiments.

\subsection{Local KL interpretation of the monitored variance}
\label{sec:kl-interpretation}

The centered log-ratio variance is interpreted as a local divergence proxy rather than as an exact KL estimator. The following second-order calculation gives its scale.

\begin{proposition}[Second-order KL proxy]
\label{prop:kl-proxy}
Let $P$ and $Q$ be two nearby completion distributions for the same prompt with common support, let $Y$ denote a completion sampled from the distribution indicated in the expectation, and define
\begin{equation*}
    \delta(y)=\log\frac{Q(y)}{P(y)}.
\end{equation*}
Assume $\delta$ is uniformly small, where $\|\delta\|_\infty$ denotes the supremum norm. Then
\begin{equation*}
\label{eq:kl-local-proxy-rigorous}
    \KL(P\|Q)
    =
    \frac{1}{2}\Var_{Y\sim P}(\delta(Y))
    +O(\|\delta\|_{\infty}^3),
\end{equation*}
and, to the same second order,
\begin{equation*}
    \KL(Q\|P)
    =
    \frac{1}{2}\Var_{Y\sim P}(\delta(Y))
    +O(\|\delta\|_{\infty}^3).
\end{equation*}
Consequently, when the current and reference policies are close and the sampled micro-batch is representative of the local completion distribution, the monitored quantity $\VarG(\delta)$ is a second-order proxy for approximately twice a local KL divergence.
\end{proposition}

Proposition~\ref{prop:kl-proxy} supports monitoring $\VarG(\delta)$ as a trust-region-like diagnostic, but the evaluated method does not use it to update $\lambda$. It does not imply that a finite sampled micro-batch variance is an unbiased or exact estimator of either forward or reverse KL. In particular, the approximation depends on the sampling distribution, the sign convention for $\delta$, and the current policy being close to the reference policy.

\section{Additional Proof Details}
\label{app:proofs}

\subsection{Proof of Lemma~\ref{lem:zero-sum-weights}}

The forward-centered surrogate is
\begin{equation*}
    \dhat_i
    =
    \delta_i
    -
    c_{\mathcal B},
    \qquad
    c_{\mathcal B}
    =
    \sg(\frac{1}{N}\sum_{j\in\mathcal B}\delta_j).
\end{equation*}
The detachment operator changes derivatives but not forward values, so in the forward pass
\begin{align*}
    \sum_{i\in\mathcal B}\dhat_i
    &=
    \sum_{i\in\mathcal B}\delta_i
    -
    \sum_{i\in\mathcal B}
    \frac{1}{N}\sum_{j\in\mathcal B}\delta_j \\
    &=
    \sum_{i\in\mathcal B}\delta_i
    -
    N\cdot
    \frac{1}{N}\sum_{j\in\mathcal B}\delta_j
    =
    0.
\end{align*}
Using $w_i=\Atilde_i-\lambda\sg(\dhat_i)$ and again using that detachment preserves forward values,
\begin{align*}
    \sum_{i\in\mathcal B}w_i
    &=
    \sum_{i\in\mathcal B}\Atilde_i
    -
    \lambda\sum_{i\in\mathcal B}\sg(\dhat_i) \\
    &=
    \sum_{i\in\mathcal B}\Atilde_i
    -
    \lambda\sum_{i\in\mathcal B}\dhat_i
    =
    \sum_{i\in\mathcal B}\Atilde_i.
\end{align*}
If the group-relative advantages sum to zero within each complete prompt group, then any micro-batch that is a union of complete prompt groups has $\sum_{i\in\mathcal B}\Atilde_i=0$. Any group-shared reward scaling preserves this zero-sum property, and the weights are zero-sum.

\subsection{Proof of Proposition~\ref{prop:rspo-feedback-gradient}}

The implemented loss is
\begin{equation*}
    \ellRSPO(\theta;\mathcal B)
    =
    -\frac{1}{N}\sum_{i\in\mathcal B}
    (\Atilde_i-\lambda\,\sg(\dhat_i))\dhat_i.
\end{equation*}
Rewards, advantages, the reference scores, $\lambda$, $\sg(\dhat_i)$, and the batch center $c_{\mathcal B}$ are constants with respect to the derivative. Since
\begin{equation*}
    \dhat_i=\delta_i-c_{\mathcal B},
    \qquad
    \nabla_\theta c_{\mathcal B}=0,
\end{equation*}
we have $\nabla_\theta\dhat_i=\nabla_\theta\delta_i$. Therefore
\begin{align*}
    \nabla_\theta\ellRSPO(\theta;\mathcal B)
    &=
    -\frac{1}{N}\sum_{i\in\mathcal B}
    (\Atilde_i-\lambda\,\sg(\dhat_i))
    \nabla_\theta\dhat_i \\
    &=
    -\frac{1}{N}\sum_{i\in\mathcal B}
    (\Atilde_i-\lambda\dhat_i)
    \nabla_\theta\delta_i(\theta),
\end{align*}
which is the stated formula.

\subsection{Proof of Corollary~\ref{cor:centered-advantage-limit}}

If the current model and reference model have identical coupled ELBO scores on the sampled completions, then
\begin{equation*}
    \widehat{\mathcal E}_{\theta}(o_i;q)
    =
    \widehat{\mathcal E}_{\mathrm{ref}}(o_i;q)
    \qquad\text{for all } i\in\mathcal B.
\end{equation*}
The per-token surrogate definition gives
\begin{equation*}
    \delta_i
    =
    \frac{
    \widehat{\mathcal E}_{\theta}(o_i;q)
    -
    \widehat{\mathcal E}_{\mathrm{ref}}(o_i;q)
    }{L_c}
    =
    0.
\end{equation*}
Therefore the detached batch center is also zero,
\begin{equation*}
    c_{\mathcal B}
    =
    \sg(\frac{1}{N}\sum_{j\in\mathcal B}\delta_j)
    =
    0,
\end{equation*}
and hence $\dhat_i=\delta_i-c_{\mathcal B}=0$ for every $i\in\mathcal B$. The RSPO weight becomes
\begin{equation*}
    w_i
    =
    \Atilde_i-\lambda\,\sg(\dhat_i)
    =
    \Atilde_i.
\end{equation*}
Using Proposition~\ref{prop:rspo-feedback-gradient},
\begin{equation*}
    \nabla_\theta\ellRSPO(\theta;\mathcal B)
    =
    -\frac{1}{N}\sum_{i\in\mathcal B}
    \Atilde_i\nabla_\theta\delta_i(\theta).
\end{equation*}
For the standard group-relative advantage-weighted surrogate,
\begin{equation*}
    \ellAW(\theta;\mathcal B)
    =
    -\frac{1}{N}\sum_{i\in\mathcal B}\Atilde_i\dhat_i(\theta),
\end{equation*}
and $\nabla_\theta\dhat_i=\nabla_\theta\delta_i$ because the center is detached. Hence
\begin{equation*}
    \nabla_\theta\ellAW(\theta;\mathcal B)
    =
    -\frac{1}{N}\sum_{i\in\mathcal B}
    \Atilde_i\nabla_\theta\delta_i(\theta)
    =
    \nabla_\theta\ellRSPO(\theta;\mathcal B).
\end{equation*}

If $\sum_{i\in\mathcal B}\Atilde_i=0$, then
\begin{align*}
    -\frac{1}{N}\sum_{i\in\mathcal B}\Atilde_i\dhat_i
    &=
    -\frac{1}{N}\sum_{i\in\mathcal B}\Atilde_i
    (\delta_i-c_{\mathcal B}) \\
    &=
    -\frac{1}{N}\sum_{i\in\mathcal B}\Atilde_i\delta_i
    +
    \frac{c_{\mathcal B}}{N}\sum_{i\in\mathcal B}\Atilde_i \\
    &=
    -\frac{1}{N}\sum_{i\in\mathcal B}\Atilde_i\delta_i.
\end{align*}
This proves the forward-value identity.

\subsection{Proof of Proposition~\ref{prop:fixed-point}}

If $\dhat_i=\Atilde_i/\lambda$ for all $i\in\mathcal B$, then
\begin{equation*}
    w_i
    =
    \Atilde_i-\lambda\sg(\dhat_i)
    =
    \Atilde_i-\lambda\dhat_i
    =
    0
\end{equation*}
for every sample in the micro-batch. Substituting $w_i=0$ into the loss gradient in Proposition~\ref{prop:rspo-feedback-gradient} gives
\begin{equation*}
    \nabla_\theta\ellRSPO(\theta;\mathcal B)
    =
    -\frac{1}{N}\sum_{i\in\mathcal B}0\cdot\nabla_\theta\delta_i(\theta)
    =
    0.
\end{equation*}
The zero-sum assumption $\sum_i\Atilde_i=0$ is needed for this target to be compatible with the forward centering identity $\sum_i\dhat_i=0$:
\begin{equation*}
    \sum_{i\in\mathcal B}\frac{\Atilde_i}{\lambda}
    =
    \frac{1}{\lambda}\sum_{i\in\mathcal B}\Atilde_i
    =
    0.
\end{equation*}

For the converse, let $J_\theta=\nabla_\theta\delta(\theta)\in\mathbb R^{N\times d}$ be the Jacobian of the micro-batch relative-score vector, where $d$ is the number of trainable parameters. Proposition~\ref{prop:rspo-feedback-gradient} can be written as
\begin{equation*}
    \nabla_\theta\ellRSPO(\theta;\mathcal B)
    =
    -\frac{1}{N}J_\theta^\top(\Atilde-\lambda\dhat).
\end{equation*}
At a stationary point this gradient is zero, so
\begin{equation*}
    J_\theta^\top(\Atilde-\lambda\dhat)=0.
\end{equation*}
If the Jacobian can realize arbitrary first-order changes in the sampled relative-score vector, equivalently if $J_\theta$ has full row rank on these sampled directions, then the nullspace of $J_\theta^\top$ is trivial. Hence
\begin{equation*}
    \Atilde-\lambda\dhat=0,
\end{equation*}
which is exactly $\dhat_i=\Atilde_i/\lambda$ for all $i\in\mathcal B$.

\subsection{Proof of Theorem~\ref{thm:first-order-equivalence-rspo}}

Recall the matched quadratic objective:
\begin{equation*}
    \ell_{\mathrm{quad},\lambda}(\theta;\mathcal B)
    =
    -\innerG{\Atilde}{\delta}
    +
    \frac{\lambda}{2}\normG{\dhat}^2,
    \qquad
    \dhat_i=\delta_i-c_{\mathcal B},
    \qquad
    \nabla_\theta c_{\mathcal B}=0.
\end{equation*}
Its gradient is
\begin{align*}
    \nabla_\theta\ell_{\mathrm{quad},\lambda}
    &=
    -\frac{1}{N}\sum_{i\in\mathcal B}
    \Atilde_i\nabla_\theta\delta_i
    +
    \frac{\lambda}{2}
    \nabla_\theta
    (
    \frac{1}{N}\sum_{i\in\mathcal B}\dhat_i^2
    ) \\
    &=
    -\frac{1}{N}\sum_{i\in\mathcal B}
    \Atilde_i\nabla_\theta\delta_i
    +
    \lambda\frac{1}{N}\sum_{i\in\mathcal B}
    \dhat_i\nabla_\theta\dhat_i \\
    &=
    -\frac{1}{N}\sum_{i\in\mathcal B}
    \Atilde_i\nabla_\theta\delta_i
    +
    \lambda\frac{1}{N}\sum_{i\in\mathcal B}
    \dhat_i\nabla_\theta\delta_i \\
    &=
    -\frac{1}{N}\sum_{i\in\mathcal B}
    (\Atilde_i-\lambda\dhat_i)\nabla_\theta\delta_i.
\end{align*}
By Proposition~\ref{prop:rspo-feedback-gradient}, this equals $\nabla_\theta\ellRSPO(\theta;\mathcal B)$. Therefore the two objectives induce the same first-order update vector for fixed $\lambda$ and fixed sampled data.

The rewriting in Eq.~\eqref{eq:quadratic-relative-score-target} follows by completing the square:
\begin{align*}
    -\innerG{\Atilde}{\delta}
    +
    \frac{\lambda}{2}\normG{\dhat}^2
    &=
    -\innerG{\Atilde}{\dhat+c_{\mathcal B}\mathbf 1}
    +
    \frac{\lambda}{2}\normG{\dhat}^2 \\
    &=
    -\innerG{\Atilde}{\dhat}
    -
    c_{\mathcal B}\innerG{\Atilde}{\mathbf 1}
    +
    \frac{\lambda}{2}\normG{\dhat}^2 \\
    &=
    \frac{\lambda}{2}
    \normG{\dhat-\frac{1}{\lambda}\Atilde}^2
    -
    \frac{1}{2\lambda}\normG{\Atilde}^2
    -
    c_{\mathcal B}\innerG{\Atilde}{\mathbf 1}.
\end{align*}

\subsection{Proof of Proposition~\ref{prop:kl-proxy}}

Because $Q$ is normalized and $\delta(y)=\log(Q(y)/P(y))$, we have
\begin{equation*}
    \E_{Y\sim P}[e^{\delta(Y)}]
    =
    \sum_y P(y)\frac{Q(y)}{P(y)}
    =
    \sum_y Q(y)
    =
    1,
\end{equation*}
with the same argument written as an integral for continuous spaces. Since $\delta$ is uniformly small, Taylor expansion gives
\begin{equation*}
    e^{\delta}
    =
    1+\delta+\frac{1}{2}\delta^2+O(\|\delta\|_\infty^3).
\end{equation*}
Taking expectation under $P$ and using $\E_P[e^\delta]=1$ yields
\begin{equation*}
    0
    =
    \E_P[\delta]
    +
    \frac{1}{2}\E_P[\delta^2]
    +
    O(\|\delta\|_\infty^3).
\end{equation*}
Thus
\begin{equation*}
    \E_P[\delta]
    =
    -\frac{1}{2}\E_P[\delta^2]
    +
    O(\|\delta\|_\infty^3).
\end{equation*}
The forward KL is
\begin{equation*}
    \KL(P\|Q)
    =
    \E_P[\log\frac{P(Y)}{Q(Y)}]
    =
    -\E_P[\delta(Y)]
    =
    \frac{1}{2}\E_P[\delta(Y)^2]
    +
    O(\|\delta\|_\infty^3).
\end{equation*}
Moreover $\E_P[\delta]=O(\|\delta\|_\infty^2)$, so
\begin{equation*}
    \Var_{Y\sim P}(\delta(Y))
    =
    \E_P[\delta(Y)^2]
    -
    \E_P[\delta(Y)]^2
    =
    \E_P[\delta(Y)^2]
    +
    O(\|\delta\|_\infty^4).
\end{equation*}
Substituting this into the previous display gives
\begin{equation*}
    \KL(P\|Q)
    =
    \frac{1}{2}\Var_{Y\sim P}(\delta(Y))
    +
    O(\|\delta\|_\infty^3).
\end{equation*}

For the reverse KL,
\begin{align*}
    \KL(Q\|P)
    &=
    \E_Q[\log\frac{Q(Y)}{P(Y)}]
    =
    \E_Q[\delta(Y)] \\
    &=
    \E_P[e^{\delta(Y)}\delta(Y)] \\
    &=
    \E_P[
    (1+\delta(Y)+O(\|\delta\|_\infty^2))\delta(Y)
    ] \\
    &=
    \E_P[\delta(Y)]
    +
    \E_P[\delta(Y)^2]
    +
    O(\|\delta\|_\infty^3) \\
    &=
    \frac{1}{2}\E_P[\delta(Y)^2]
    +
    O(\|\delta\|_\infty^3) \\
    &=
    \frac{1}{2}\Var_{Y\sim P}(\delta(Y))
    +
    O(\|\delta\|_\infty^3).
\end{align*}
This proves both claims.

\subsection{Perturbation from surrogate error}
\label{app:delta-error-new}

The theory above treats $\delta_i$ as a log-ratio surrogate. Let $R_i$ denote an ideal uncentered log-ratio value for sample $i$. Suppose the corresponding ideal centered relative-score vector on the current micro-batch is $\widehat R_i=R_i-N^{-1}\sum_{j\in\mathcal B}R_j$, and the implemented centered relative-score surrogate is
\begin{equation*}
    \dhat_i=\widehat R_i+\widehat\xi_i,
\end{equation*}
where $\xi_i$ is the uncentered surrogate error, $\widehat\xi_i$ is the same error after centering, and $\epsilon$ is a uniform error bound satisfying $|\xi_i|\le \epsilon$ before centering. Then $|\widehat\xi_i|\le 2\epsilon$ and $\normG{\widehat\xi}\le 2\epsilon$.

Let $\ellAW^\delta$ and $\ellAW^R$ denote the standard group-relative advantage-weighted surrogate evaluated with the implemented surrogate $\dhat$ and the ideal centered relative score $\widehat R$, respectively. Define $\ellRSPO^\delta$ and $\ellRSPO^R$ analogously for the RSPO objective.

For the standard group-relative advantage-weighted surrogate,
\begin{equation*}
    |\ellAW^\delta-\ellAW^R|
    =
    |\innerG{\Atilde}{\widehat\xi}|
    \le
    \normG{\Atilde}\,\normG{\widehat\xi}
    \le
    2\epsilon\normG{\Atilde}.
\end{equation*}
For the variance term,
\begin{align*}
    |\normG{\dhat}^2-\normG{\widehat R}^2|
    &\le
    2\normG{\widehat R}\normG{\widehat\xi}
    +
    \normG{\widehat\xi}^2 \\
    &\le
    4\epsilon\normG{\widehat R}+4\epsilon^2.
\end{align*}
Consequently, for a fixed $\lambda$, the forward RSPO loss values satisfy
\begin{equation*}
    |\ellRSPO^\delta-\ellRSPO^R|
    \le
    2\epsilon\normG{\Atilde}
    +
    \lambda(4\epsilon\normG{\widehat R}+4\epsilon^2).
\end{equation*}
This bound controls the scalar surrogate error. Gradient error additionally depends on the Jacobian error of the ELBO estimator and is therefore implementation-dependent.

\section{Implementation Notes}
\label{app:implementation-notes}

\paragraph{Advantage scaling.}
RSPO uses zero-sum group-relative advantages. Here $s_q$ denotes the prompt-level advantage scale and $\sigma_r$ denotes the reward standard deviation within the prompt group. The main experiments use the unscaled advantage $s_q=1$, while the normalized ablation uses $s_q=\sigma_r+10^{-4}$. When reward scaling is enabled, the sample standard deviation is computed using the training framework's standard convention.

\paragraph{Fixed feedback coefficient and inactive components.}
The value $\lambda$ used in every backward pass is fixed rather than adaptively updated; the main experiments use $\lambda=0.01$, while ablations also test other fixed values. Adaptive updates of $\lambda$, reference-model exponential-moving-average (EMA) updates, and removal of zero-standard-deviation reward groups are not used in the reported experiments.

\paragraph{Detached batch centering.}
The batch mean subtracted from $\delta$ is detached before subtraction. This makes the forward values centered but leaves $\nabla_\theta\dhat_i=\nabla_\theta\delta_i$ in the backward pass.

\paragraph{Reference-model coupling.}
For each completion, the current and reference scores share the same mask time and mask set. Without this coupling, the difference $\delta_i$ includes unnecessary independent Monte Carlo noise.

\paragraph{Zero-variance reward groups.}
If the group reward standard deviation is numerically zero, then all rewards in the group are equal and the group has no relative preference signal. Such groups are retained in the reported training runs; the zero-standard-deviation ratio is recorded only as a diagnostic.

\section{Additional Experimental Details}
\label{app:experimental-details}


All reinforcement learning training is implemented with the TRL library~\citep{vonwerra2020trl}. 
We conduct experiments on mathematical reasoning and planning benchmarks, including GSM8K, Math500, Countdown, and Sudoku. All experiments were conducted on 8 NVIDIA H100-80G GPUs

\begin{figure}[t]
    \centering
    \includegraphics[width=0.48\linewidth]{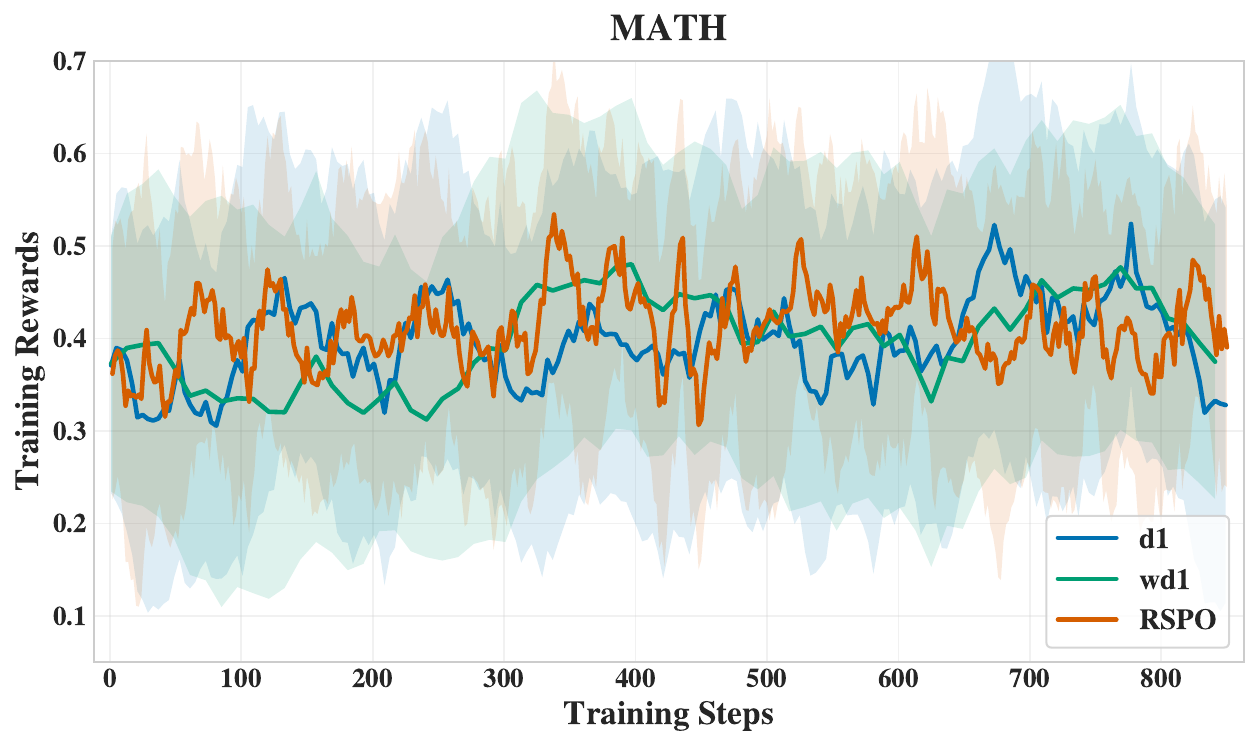}
    \hfill
    \includegraphics[width=0.48\linewidth]{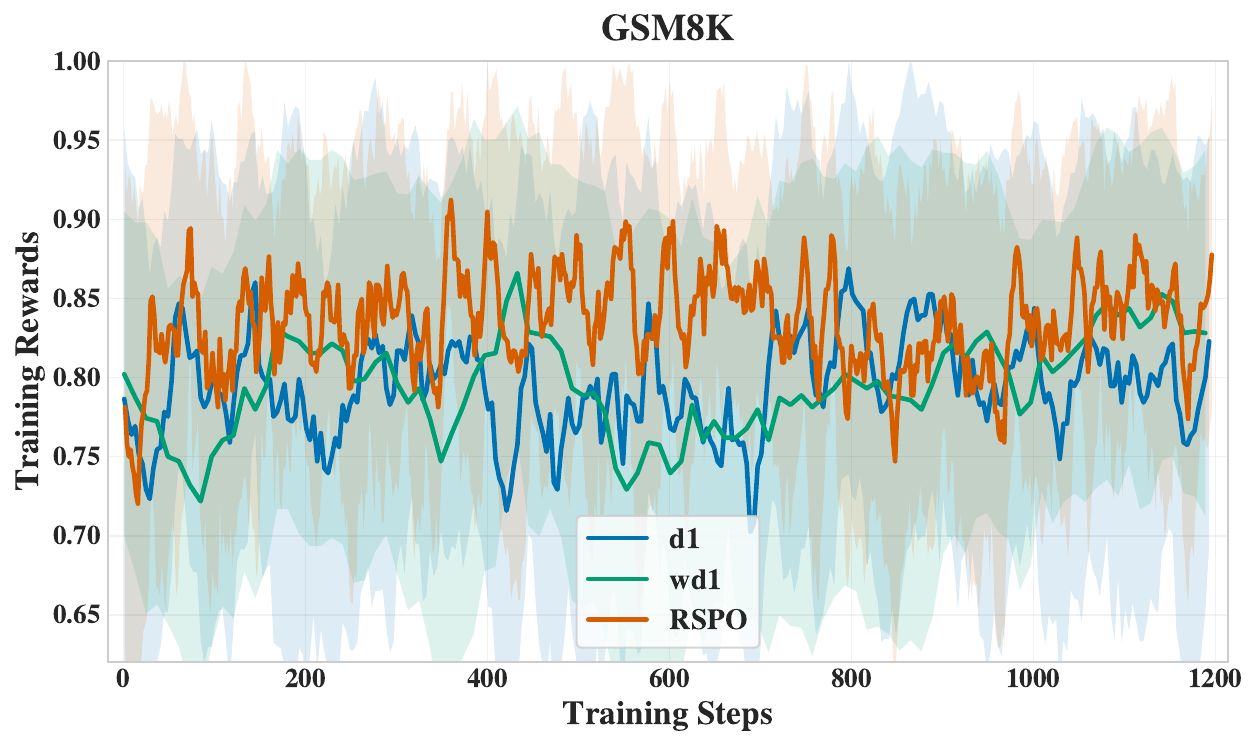}
    \caption{Additional training reward dynamics on mathematical reasoning benchmarks. Shaded regions indicate variation across runs.}
    \label{fig:reward-dynamics-math}
\end{figure}

\paragraph{Parameter-Efficient Fine-Tuning.}
For all experiments, we apply Low-Rank Adaptation (LoRA)~\citep{hu2022lora} to the base diffusion language model. 
The LoRA rank is set to $r=128$ and the scaling factor is set to $\alpha=64$. 
All reported results are obtained with LoRA fine-tuning rather than full-parameter fine-tuning.

\paragraph{Optimization.}
We optimize the policy with AdamW~\citep{loshchilov2017decoupled} using $\beta_1=0.9$ and $\beta_2=0.99$. 
A constant learning-rate schedule is used throughout training. 
Unless otherwise specified, the learning rate is set to $3\times 10^{-6}$, weight decay to $0.01$, and gradient clipping to $0.2$. 
For estimating ELBO-based relative scores, we use $K=2$ Monte Carlo samples for computational efficiency. 

\paragraph{Batching.}
We set the group size according to task difficulty. 
For GSM8K, Countdown, and Sudoku, we use group size $G=6$ and total batch size $96$. 
For MATH500, we use group size $G=16$ and total batch size $256$. 
Gradient accumulation is applied to reach the target effective batch size.

\paragraph{Training Steps and Checkpoint Selection.}
Models are trained until the reward curve stabilizes. 
For mathematical reasoning, we train GSM8K and MATH500 for up to $1.5$k steps. 
For planning tasks, we train models for up to $2$k steps. 
During training, checkpoints are evaluated periodically, and we report the checkpoint with the highest average test accuracy across generation lengths $256$ and $512$.

\paragraph{Decoding and Evaluation.}
All experiments are conducted in the zero-shot setting. 
For both RL rollouts and evaluation, we use semi-autoregressive confidence-based decoding following prior dLLM work. 
The denoising timestep is set to half of the total generation length, and the sequence is divided into blocks of $32$ tokens. 
At each diffusion step, the model unmasks the two tokens with the highest confidence within the current incomplete block. 
During RL rollouts, the generation length is set to $256$ for all tasks. 
The sampling temperature is set to $0.9$ for GSM8K, MATH500, and Countdown, and to $0.3$ for Sudoku. 
During evaluation, the temperature is set to $0$.

\paragraph{Special Notes on Sudoku.}
Sudoku evaluation follows the verifier-based setting used in prior dLLM planning work. 
A generated solution is considered correct only if it satisfies the Sudoku constraints and is consistent with the given puzzle. 
Although some prior work uses few-shot prompting for Sudoku, our reported experiments use zero-shot prompting consistently for both training and evaluation.

\end{document}